\documentclass[10pt,twocolumn,letterpaper]{article}

\usepackage{iccv}
\usepackage{times}
\usepackage{epsfig}
\usepackage{graphicx}
\usepackage{amsmath}
\usepackage{amssymb}

\usepackage{comment}
\usepackage{multirow}
\usepackage[table,xcdraw]{xcolor}

\usepackage[pagebackref=true,breaklinks=true,letterpaper=true,colorlinks,bookmarks=false]{hyperref}

\usepackage{capt-of}
\usepackage{soul}
\usepackage{wrapfig}
\usepackage{booktabs}
\usepackage{float}
\usepackage{diagbox}
\def\iccvPaperID{2062} % *** Enter the ICCV Paper ID here

\iccvfinalcopy
\ificcvfinal\fi

\newcommand{\mysubsubsection}[1]{\vspace{0.1cm} \noindent {\bf #1}.}
\newcommand{\mysubsubsectiona}[1]{\vspace{0.1cm} \noindent {\bf #1}}
\newcommand{\myetal}[0]{\textit{et al}.~}

\begin{document}

%%%%%%%%% TITLE
\title{Structured Outdoor Architecture Reconstruction\\ by Exploration and Classification}
%\title{Search and Evaluate for Building Structure Estimation}

\author{Fuyang Zhang \qquad Xiang Xu \qquad Nelson Nauata \qquad Yasutaka Furukawa\\
Simon Fraser University, BC, Canada\\
{\tt\small \{fuyangz, xuxiangx, nnauata, furukawa\}@sfu.ca}
% For a paper whose authors are all at the same institution,
% omit the following lines up until the closing ``}''.
% Additional authors and addresses can be added with ``\and'',
% just like the second author.
% To save space, use either the email address or home page, not both
}

%\maketitle
% Remove page # from the first page of camera-ready.
\ificcvfinal\thispagestyle{empty}\fi

\twocolumn[{
 \maketitle
 \vspace{-2em}
 \centerline{
 \includegraphics[width=\textwidth]{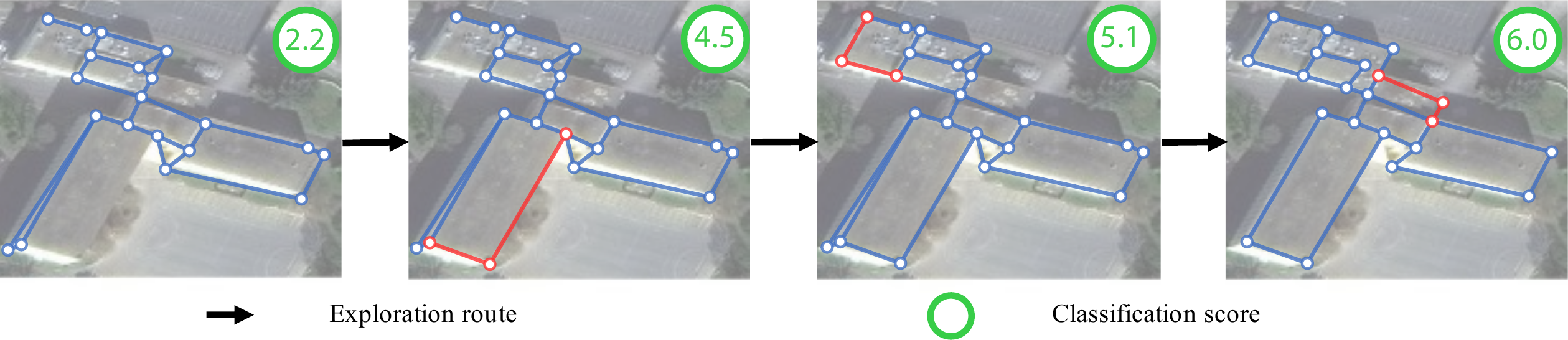} 
}
\captionof{figure}{Our explore-and-classify reconstruction framework iterates between exploring new building structures by heuristic actions and classifying the correctness of geometries. The figure shows a particular path of exploration by our system. The red color highlights the major improvements against the previous step.
%This example demonstrates 4 iterations between exploring new building structures and keeping the result with the highest graph score. Result after our explore-and-classify framework correctly reconstructs majority of the building structure.  
}
\label{fig:teaser}
 \vspace{1em}
}]

%%%%%%%%% ABSTRACT
\begin{abstract}
   This paper presents an explore-and-classify framework for structured architectural reconstruction from an aerial image.
%, which is notoriously difficult.
%it is more effective to learn the task of classifying a given reconstruction to be "Good/Bad", instead of the task of reconstructing the best structured geometry mode, which is notoriously difficult.
%
%recovering structured outdoor architecture from aerial images. 
Starting from a potentially imperfect building reconstruction by an existing algorithm, 
%initial reconstruction graph; 
our approach 1) explores the space of building models by modifying the reconstruction via heuristic actions; 2) learns to classify the correctness of building models while generating classification labels based on the ground-truth; and 3) repeat.
%generates classification labels for new samples by using the ground-truth; 3) learns to rank them by classifying the correctness of local geometries; and 4) repeat.
%geometric primitives within.
%
At test time, we iterate exploration and classification, seeking for a result with the best classification score.
%the searched graph with the best overall score is our final solution.
We evaluate the approach using initial reconstructions by two baselines and two state-of-the-art reconstruction algorithms.
%initial graph from two baselines and two state-of-the-art algorithms.
Qualitative and quantitative evaluations demonstrate that our approach consistently improves the reconstruction quality from every initial reconstruction.
%starting from any of the existing results. 
%We further analyze the connection to reinforcement learning (RL) and show our superiority over two state-of-the-art RL-based reconstruction systems. 

%while preparing the classificaiton labels

%taking actions to evolve geometric models (exploration) and scoring the goodness of the models (classification)

%1) exploring the solution space and 2) ranking the models by classifying the goodness of geometric primitives

%which repeatedly explore and classify geometric models 

%. The main idea to consider reconstruction process as an iterative exploration and classification

%. , which    (explore) (classify)

%TODO

%use \textbf{$\backslash$fuyang\{blabla\}}(\fuyang{blabla}), to write comments. \textbf{$\backslash$sam\{...\}}, \textbf{$\backslash$yasu\{...\}}
\end{abstract}

%%%%%%%%% BODY TEXT
\section{Introduction}

Imagine a task of drawing to reconstruct the face of a person you met in the morning. This is a challenging task,
%where it is difficult to produce a reconstruction in one shot. 
where most people would perform trial and error, that is, iterate modifying and verifying the face reconstruction until being satisfied with the result.
%until satisfaction.
%correctness of the modifications. 
%
%exploring the drawing by modifying parts and 
%An easier way is to simply answer Yes/No to face-reconstructions made by a trained artist, who refines the facial drawing based on the answers.
In the space of structured reconstruction, which is notoriously difficult even with the emergence of deep learning, researchers compete on designing effective neural architectures that directly reconstructs structured geometry (or its constituent information)~\cite{liu2017raster,zhang2020conv,yu2019single,zhou2019wireframe}. Our idea is to decompose the process of reconstruction into the iteration of trial and error, where the core learning is in verifying the correctness of a given geometry as opposed to one-shot geometry regression.

More concretely, the paper proposes a novel explore-and-classify approach for structured reconstruction. Starting from an imperfect building reconstruction by an existing algorithm, 
our approach 
uses simple geometry-editing actions to explore a set of offspring graphs and learns to classify the correctness of them at different levels of local geometries. 
%1) explores the space of building models by modifying the reconstruction via heuristic geometry-editing actions; 2) learns to classify the correctness of buildings models at three levels of local geometries, while generating correct/incorrect classification labels based on the ground-truth; and 3) repeat. 
At test time, we iterate exploration and classification with a search method such as beam-search or sequential Monte Carlo \cite{doucet2001smc}, trying to find the result with the best classification score.

We demonstrate the effectiveness of the approach on the task of structured outdoor architecture reconstruction from an aerial image~\cite{nauata2020vectorizing}, where a building structure is represented as a graph (See Fig.~\ref{fig:teaser}). We use two baselines and two state-of-the-art algorithms~\cite{zhang2020conv,nauata2020vectorizing} to generate the initial reconstructions. Our system consistently makes big improvements over all the initial reconstructions and achieves state-of-the-art performance. 
%\fuyang{Remove following sentence?}
%We also provide in-depth analysis on the connection to deep reinforcement learning (DRL), and conduct ablation studies showing our superiority over two state-of-the-art DRL based reconstruction systems~\cite{ellis2019write,lin2020modeling}.

In summary, the contributions of this paper are two-fold: 1) A novel explore-and-classify framework with local geometry classifiers for a structured reconstruction problem; and
%; 2) A local geometry classifier, which allows us to 
%\sam{(2) use of local-primitive supervision}and 
2) A state-of-the-art performance on outdoor architecture reconstruction problem with significant improvements over all the baselines and the best existing approaches.
%we cast the problem of reconstruction as a graph classification problem, (2) we show the local-primitive supervision is crucial for structured reconstruction problem and (3) we present a novel system that combines heuristic policy, classification network as well as searching scheme to improve the reconstruction. 
%
While we focus on one problem setting, the ideas are potentially applicable to many other structured reconstruction tasks and could further improve the performance of their state-of-the-arts.
Our code and pretrained models are available at \url{https://github.com/zhangfuyang/search_evaluate}.
The data is publicly available through our prior project~\cite{nauata2020vectorizing}.

\section{Related Work}

In this section, we review related work in the field of structured geometry reconstruction.

\mysubsubsection{Traditional methods}
%Structured reconstruction has a long history in computer vision.
%
Birchfield and Tomasi had a seminal paper on piecewise affine surface reconstruction, which iterates plane segmentation and affine parameter estimation~\cite{birchfield1999multiway}. The development of graphical model inference techniques such as graph-cuts~\cite{kolmogorov2002multi}, alpha-expansion~\cite{kolmogorov2004energy}, and message passing~\cite{kolmogorov2006convergent} enabled effective optimization algorithms for piecewise smooth depthmap reconstruction~\cite{furukawa2009manhattan,gallup2010piecewise}, which produces a consistent 3D planes under the Manhattan constraint. However, they were not robust enough for production applications due to the use of many heuristics and unstable depth estimation.

\mysubsubsection{Arrival of deep learning}
The emergence of deep neural networks (DNNs) has brought many breakthroughs. Instance segmentation networks~\cite{he2017mask} in recognition were utilized for single-view piecewise planar depthmap reconstruction~\cite{liu2018planenet}. A metric learning approach for instance segmentation has also been proven effective for the same task~\cite{yu2019single}. Recurrent neural networks~\cite{castrejon2017polygon,acuna2018polygonrnn++} or iterative refinement approaches are utilized for polygonal curve extraction~\cite{cheng2019darnet}. However, these approaches are segmentation methods where the region boundary is a dense set of points.

For reconstruction of CAD-level geometry with compact surface representation, a popular approach is to utilize DNNs for low level geometry inference (e.g., corner or edge detection) and optimization-based methods use heuristics to reconstruct high level topological structure~\cite{liu2017raster,chen2019floor,nauata2020vectorizing}.

End-to-end neural architecture for structured geometry reconstruction also exists, albeit more challenging. Convolutional graph neural networks are proposed for a 2D planar graph reconstruction given corners~\cite{zhang2020conv}. For wire-frame parsing of architectural scenes, end-to-end systems combine junction detection and edge verification~\cite{zhou2019wireframe} or junction detection and adjacency matrix inference~\cite{zhang2019ppgnet}.

Despite continued progress in neural architecture research, the task of CAD-quality geometry reconstruction is still a challenge, where state-of-the-art algorithms produce reconstructions that immediately look ``wrong'' to our eyes.

%Junction detection and edge verification is trained end-to-end for wire-frame parsing of architectural scenes~\cite{zhou2019wireframe}. Similarly, junction detection and the adjacency matrix inference is trained end-to-end for wire-frame parsing~\cite{zhang2019ppgnet}.

\mysubsubsection{Reinforcement learning}
Deep reinforcement learning (DRL) is an interesting way to tackle structured geometry reconstruction.
Ellis~\myetal\cite{ellis2019write} proposed a neural program synthesis algorithm, one of whose applications is constructive solid geometry reconstruction from a 2D or 3D occupancy grid. Lin \myetal\cite{lin2020modeling} proposed a system that takes a depthmap and recovers a set of primitives and refines corner locations. DRL sounds elegant but requires many tricks and heuristics, notably pseudo supervised training data generation by domain-specific knowledge.
%network supervised pre-training where domain-specific heuristics are used to generate supervised training data. 
Sec.~\ref{rl_setup} explains that our system is a special case of DRL in a simpler form while standard DRL features are not effective
%and what aspects of DRL is important 
for structured geometry reconstruction.
%without tricks or certain system components.
Our experiments show that our system makes significant improvements over the state-of-the-art DRL based reconstruction methods.

%the task of geometric level reasoning, e.g. the CAD-quality building reconstruction, is far from product-level performance, which still remains as a challenge. The current state-of-the-art method \cite{zhang2020conv, zhang2019ppgnet, zhou2019wireframe, huang2018learning} tries to utilize a fully deep learning method to solve the task, which successes in primitives detection and makes progress for general structure understanding but fails in local topology reasoning.

%\fuyang{methods that directly reconstruct graph}

%Computer vision is still at infant stage in holistic reasoning of geometric structure. For visual perception as well as detection, deep neural networks have been effective models after decades of exploration. However, 

%\textbf{learning:} 

%conv-mpn \cite{zhang2020conv}, 

%wireframe: PPGNet \cite{zhang2019ppgnet}, L-CNN \cite{zhou2019wireframe}, \cite{huang2018learning},

%polygonal loop: 

%polygon-rnn \cite{castrejon2017polygon}, polygon-rnn++ \cite{acuna2018polygonrnn++}, darnet \cite{cheng2019darnet}, active contour \cite{marcos2018activecontour}

%Zeng~\etal \cite{zeng2018neural}

%\textbf{optimization:}

%nauata~\etal \cite{nauata2020vectorizing},
%floor-sp \cite{chen2019floor},
%floornet \cite{liu2018floornet},
%\cite{liu2017raster}

%planercnn, associative embedding by shanghai tech

%\fuyang{methods that use value function}
%
%REPL \cite{ellis2019write}, TUM \cite{lin2020modeling}
\section{Problem review}

%%% I heavily optimize this paragraph witha wrapfigure. Very careful in editing even a single word.
The paper tackles a structured architecture reconstruction problem by Nauata \myetal\cite{nauata2020vectorizing}. 
%Nauata \myetal proposed a structured architecture reconstruction problem~\cite{nauata2020vectorizing}.
%An input is 
An input is a cropped and rescaled (i.e., 256$\times$256) aerial photograph of a building.
%with a resolution of 256$\times$256.
%
The task is to reconstruct the building architecture as a 2D planar graph including all internal feature edges. 
\begin{wrapfigure}[5]{r}{1.9cm}
\vspace{-0.52cm}
\hspace{-0.5cm}
\includegraphics[width=2.2cm]{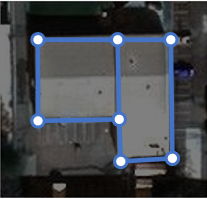}
\end{wrapfigure}
Suppose there exists an L shaped building consisting of two rectangular components (See the right). The output must have two T-junctions and five L-junctions with the shared internal edge.

We use the same metrics for evaluation as in Nauata \myetal\cite{nauata2020vectorizing}, namely
%We evaluate the quality of reconstructed planar graphs by
f1 scores of corner, edge, and region
primitives. 
The dataset contains 2,001 buildings. We randomly split the data into 1601 training, 50 validation, and 350 testing data. 
Hyper-parameters such as score weights, search depth and search width are fine-tuned on the validation set. We report all result on the test set.
Note that an aerial photograph suffers from perspective distortions, where the Manhattan assumption does not always hold, yielding a very challenging structured reconstruction problem.

\section{Explore-and-classify reconstruction}
%by Exploration and Classification}

\begin{figure*}
    \centering
    \includegraphics[width=\textwidth]{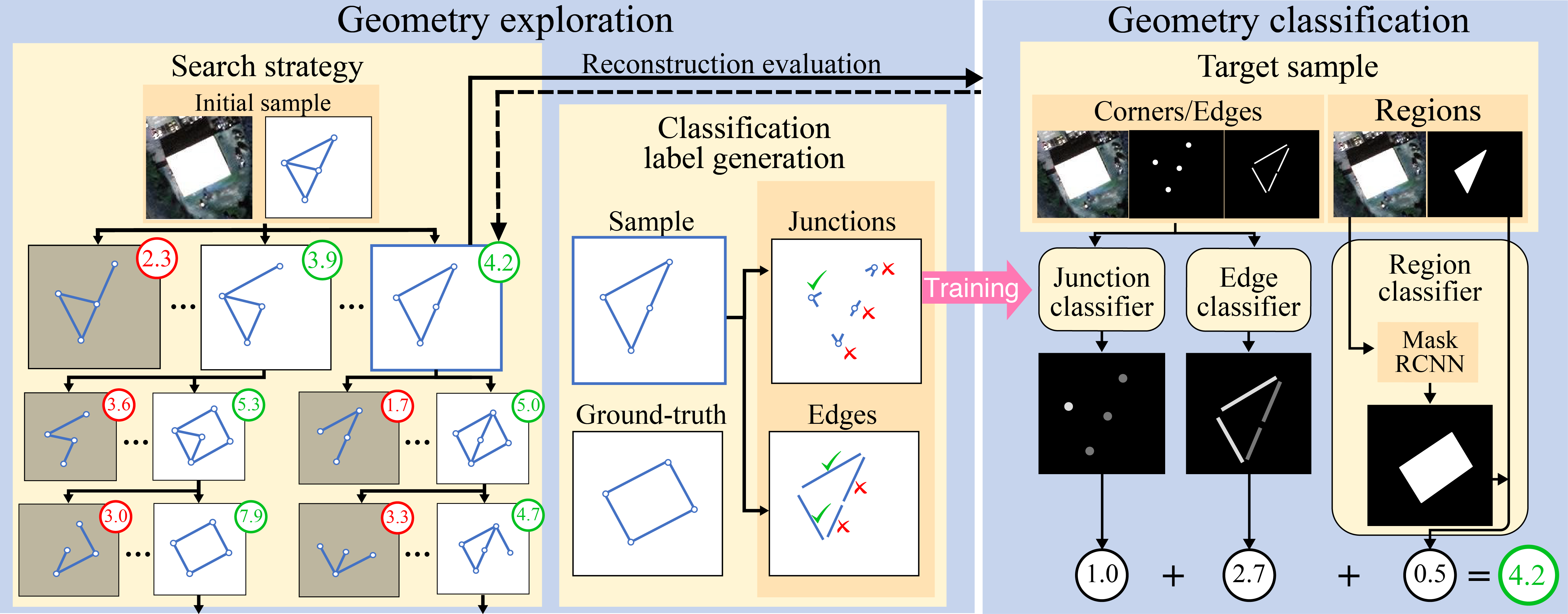}
    %\vspace{-0.4cm}
    \caption{System overview. The geometry exploration module starts from an initial model reconstructed by another algorithm, and produces a set of offspring reconstructions by heuristic actions. The geometry classification module ranks the generated reconstructions and keep the top samples for the next iteration (graphs with green color score on the right corner are the top samples in each iteration).
    %
    %First, geometry exploration. Given an reconstruction graph, geometry exploration module produces a set of offspring-graphs by applying heuristic actions. The geometry classification module is used to evaluate all of them and keep top-k for next iteration. 
    %
    The system dynamically generates training data for the classification module by picking generated reconstructions and creating classification labels based on the ground-truth.
    The classification module consists of three local primitive classifiers (i.e., junctions, edges, and regions).
     The first two classifiers are neural networks, which learns to classify if each primitive in the current reconstruction is correct or incorrect. The region classifier measures the consistency with the instance segmentation masks generated by Mask R-CNN~\cite{he2017mask}.
    %evaluates the difference between the current 
    The overall classification score is the sum of junction, edge, and region scores.
    }
    \label{fig:overview}
\end{figure*}

% \subsection{System overview}
%The fundamental idea of this paper is simple yet powerful. Standard methods reconstruct graphs from the input image as one-shot geometry regression, from which is extremely hard to infer primitives relationships. 
Our idea is to decompose the process of reconstruction into the iteration of exploration and classification, where the core learning happens in classifying the correctness of a geometry (see Fig.~\ref{fig:overview}). 
%Our system is similar in spirit to reinforcement learning (RL) with a value    network~\cite{ellis2019write,lin2020modeling}, where Sec.~\ref{rl_setup} provides a through comparative analysis to exalt our contributions over standard RL methods.
%, we dedicate Section~\ref{section:rl} for a through comparative analysis. 
%After reviewing the problem definition, 
%This section explains the exploration and classification modules in our system.

\subsection{Geometry exploration} \label{sec:exploration}
%\mysubsubsection{Search strategy}
The exploration module takes a building graph and produces a set of offspring-graphs by heuristic actions.
%\sam{why a set of building graphs? I thought input is always the initial result from existing algorithm} 
We use the geometry classification module to rank the offspring, shrink the population (e.g., keeping top-k), and repeat the iteration of exploration and classification. The exact procedure slightly differs during testing and training. We now explain
1) the heuristic actions, 2) the test-time exploration, and 3) the training-time exploration.

\begin{figure}
    \centering
    \includegraphics[width=\columnwidth]{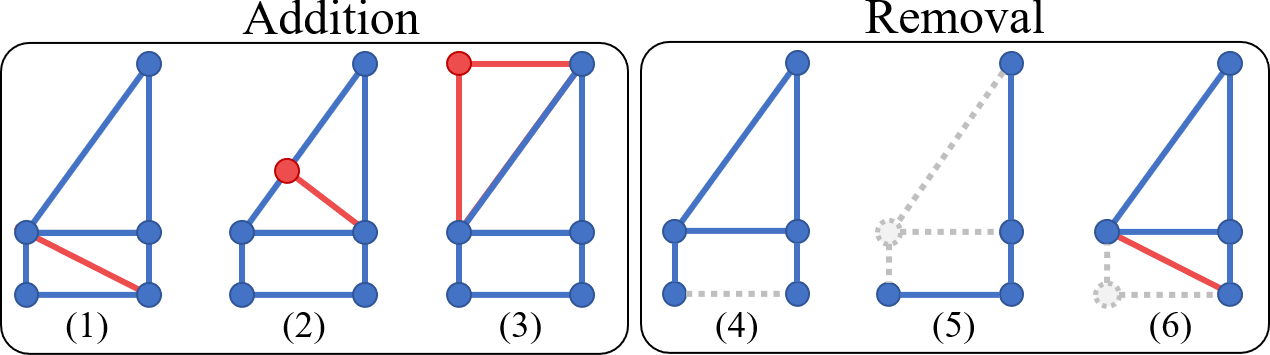}
    %\vspace{-0.4cm}
    \caption{Heuristic actions for geometry modification. Red and grey colors indicate the added and deleted elements respectively.}
    %,  adding a new edge, red circle is adding a new corner. Grey dashed line or circle is removing an edge or corner.}
    \label{fig:actions}
\end{figure}

\mysubsubsection{Heuristic actions}
The heuristic actions either add primitives or remove primitives (See Fig.~\ref{fig:actions}). There are three addition actions: 1) Add an edge between a pair of existing corners; 2) Add an edge starting from a corner orthogonally towards another edge and insert a new corner at the intersection (if the intersection is outside, we extend the edge until the intersection point);
%At one existing corner, add an orthogonal edge of an existing edge, and insert a new corner at the intersection place; 
3) At two edges sharing a corner, complete a parallelogram by adding one corner and two edges. 
There are three removal actions: 4) Remove an edge; 5) Remove a corner and its incident edges; 6) For a degree-2 corner, remove itself and connect its end-points, which is effective for removing a colinear corner.
%There are three removal actions: 1) remove a corner and its incident edges, 2) remove a single edge and 3) remove two colinear edges and connect its endpoints. These actions are designed to narrow the search space down to primitives connected to the current graph. 

%To ensure graph planarity, an action is not executed when the new edge creates a self-intersection. The planarity can be learned by our classifiers, but this ad-hoc filter is introduced in our system for efficiency.

\mysubsubsection{Test-time exploration}
Given a set of building graphs, we apply every possible action to every graph, evaluate the scores based on the classifiers (See Sec.~\ref{sec:classifiers}), and subsample the population
%During inference phase, our method favors reconstruction accuracy over runtime performance. We apply the entire set of available actions, assess the entire offspring and subsample them 
using Beam search or Sequential Monte Carlo (SMC)~\cite{doucet2001smc}. The process repeats for a certain number of times and the graph with the highest classification score becomes the output. %\yasu{Some hyper params of beam search and smc if not too long here.} 
Our default choice is the Beam search with (depth=12, width=5), while we increase the depth to 20 or 30 depending on the quality of the initial reconstructions. See Sec.~\ref{sec:results} for details.

\mysubsubsection{Training-time exploration}
During training, there is no need to seek for the best reconstruction for each sample, and we favor population diversity and efficiency. We take one action for each of the 6 action types for each graph. For example, an edge removal action is applied to one edge with a uniform probability. In each iteration, we keep the top two graphs based on the classification scores, while using an epsilon-greedy strategy to replace one with a random graph with 20\% chance.
%We also use an epsilon-greedy strategy where there is a small probability (0.2) for randomly choosing a graph to replace the one in top-2. 

Overall, starting from each training building, we generate 12(=6$\times$2) candidate graphs per iteration (6 in the first iteration), and repeat for 5 times.
%(search depth=5).
%
The top 2 graphs (w/ random replacement) at each of the five iterations, that is, 10 building graphs are added to the training set. See Sec.~\ref{sec:classifiers} for the classification label generation.

\begin{figure}
    \centering
    \includegraphics[width=\linewidth]{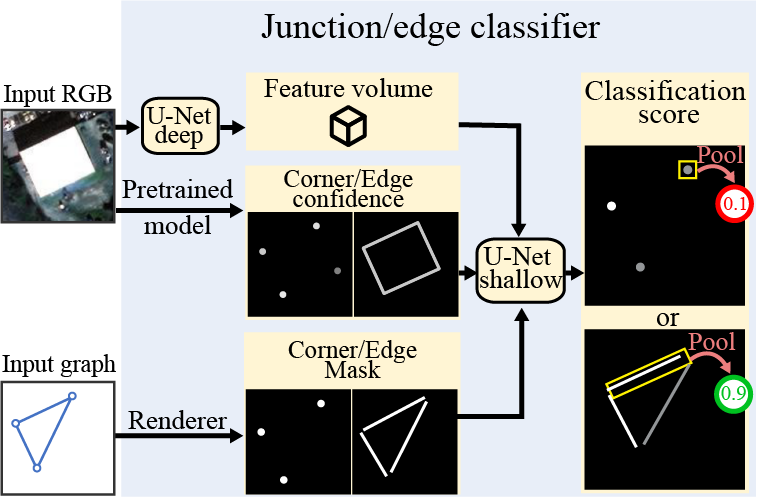}
    %\vspace{-0.4cm}
    \caption{Junction/edge classifier. The architecture consists
    %U-Net is the backbone architecture. 
    %the classifier is consist 
    of two U-Net models (U-Net deep and U-Net shallow). U-Net deep transforms an input image into a feature volume. 
    %encodes general image feature, while
    U-Net shallow takes the feature volume, a corner/edge confidence image generated by a pre-trained CNN, and a corner/edge mask image rasterized from the current graph as the input. It then produces a pixel-wise classification score. We use average pooling among pixels over a primitive to compute the final classification score for each primitive.%predicts a pixel-wise correct/incorrect classification image. Then we use average pool among the each primitive to compute the classification score.
    }
    \label{fig:network_structure}
\end{figure}

\subsection{Geometry classification} \label{classification}
\label{sec:classifiers}
We define three levels of geometric primitives in a building graph, namely junctions $\mathcal{J}$=$\{j\}$, edges $\mathcal{E}$=$\{e\}$, and regions $\mathcal{R}$=$\{r\}$. Note that a junction is a corner with information of the incident edge angles and a region is a 2D polygon surrounded by a set of edges. Our geometry classification learns to classify the correctness of each geometric primitives. The classification score is:
\begin{equation}\label{eq:classifier_score} 
w_{j}
\underset{j \in \mathcal{J}}{\sum} C_{\text{junc}}(j) + 
w_{e}
\underset{e \in \mathcal{E}}{\sum} C_{\text{edge}}(e) +
w_{r} 
C_{\text{region}}(\underset{r \in \mathcal{R}}{\cup} r).
\end{equation}
\noindent 
%$w_{\hat{j}}, w_{\hat{e}}, w_{\hat{r}}$ are the scaling weights (see Sec.~\ref{sec:results} for the values).
$C_{\text{junc}}(j)$ and $C_{\text{edge}}(e)$ are the junction/edge classification scores.  $C_{\text{region}}(r)$ is the region score. The range of $C_{\text{junc}}$, $C_{\text{edge}}$ and $C_{\text{region}}$ are $[-1,1]$, $[-1,1]$, and $[0,1]$ respectively. Junction and edge classification scores are summed over the primitive instances. 
%the junction and edge score are $[-|\mathcal{J}|, |\mathcal{J}|]$ and $[-|\mathcal{E}|, |\mathcal{E}|]$, where $|\mathcal{J}|$ and $|\mathcal{E}|$ are the number of junctions and edges respectively.
Region score is evaluated for the union of all the region instances. $w_{j,e,r}$ are the scaling weights.  We now go over the details.  

\mysubsubsection{Junction/edge classifiers}
We use the same architecture for the junction and edge classifiers. Their inputs are the aerial image and a building graph, and the output is pixel-wise classification scores. % within the range [-1 1].  

%Fig.~\ref{fig:network_structure} shows an overview of our junction/edge classifiers, 
The classifier consists of two U-Nets (See Fig.~\ref{fig:network_structure}).
%which can be divided into two parts: a deep U-Net and another shallow.
The first ``deep U-Net'' is responsible for extracting global building feature volumes. At test time, deep U-Net runs only once for each building image, significantly reducing the computational expenses.
%, which can be computed only once and shared for the same input building image during test time. This significantly reduces the running time at test time. 
The second ``shallow U-Net'' (half resolution and 1 less layer) is responsible for producing the pixel-wise classification scores as an image. The input is the feature volume, corner/edge confidence images as in Nauata~\etal~\cite{nauata2020vectorizing}, and the corner/edge binary segmentation masks, rasterized from the current building graph (3 pixels of diameters for corners and 2 pixels of thickness for edges).

We take the score at a corner pixel as the corner classification score $C_{\text{junc}}$, and take the average over the pixels along an edge as the edge classification score $C_{\text{edge}}$.
See the supplementary document for the full architecture details. 
%
%is responsible for assigning the final primitive scores using the global feature volumes plus, corner/edge confidence images and corner/edge masks from the graph, as shown in Fig.~\ref{fig:network_structure}. 
%We also use a pretrained model to extract corner/edge confidence images as in Nauata~\etal~\cite{nauata2020vectorizing}. The corner/edge masks are obtained by rendering corners and edges with 3 pixels of diameters and 2 pixels of thickness, respectively. The shallow U-Net outputs a pixel-wise classification score, 
%which are average-pooled over primitive locations for obtaining the final classification score $C_{\text{junc}}(j)$, $C_{\text{edge}}(e)$ for junctions and edges

%%%%%%%%%%%%%%%%%%%%%%%%%%%%%%%%%%%%%%%%%%%%%%%%%%%%%%%
\mysubsubsection{Region classifier}
A natural extension of the junction/edge classifiers to region is to label each region instance as correct/incorrect and train a similar classifier. 
However, such an approach degraded performance.
We found that this is due to the ambiguity in the annotations.
%local regions is subjective to annotation ambiguity. 
Suppose a system reconstructs a building with two regions in the image below, but the annotation is a single region (w/o the orange structure). All the regions will have small IoU against ground-truth and become incorrect. On the other hand, influences to the junction/edge classifiers are smaller: only 2 out of 9 junctions and
\begin{wrapfigure}[7]{r}{2.6cm}
\vspace{-0.45cm}
\hspace{-0.5cm}
\includegraphics[width=3.0cm]{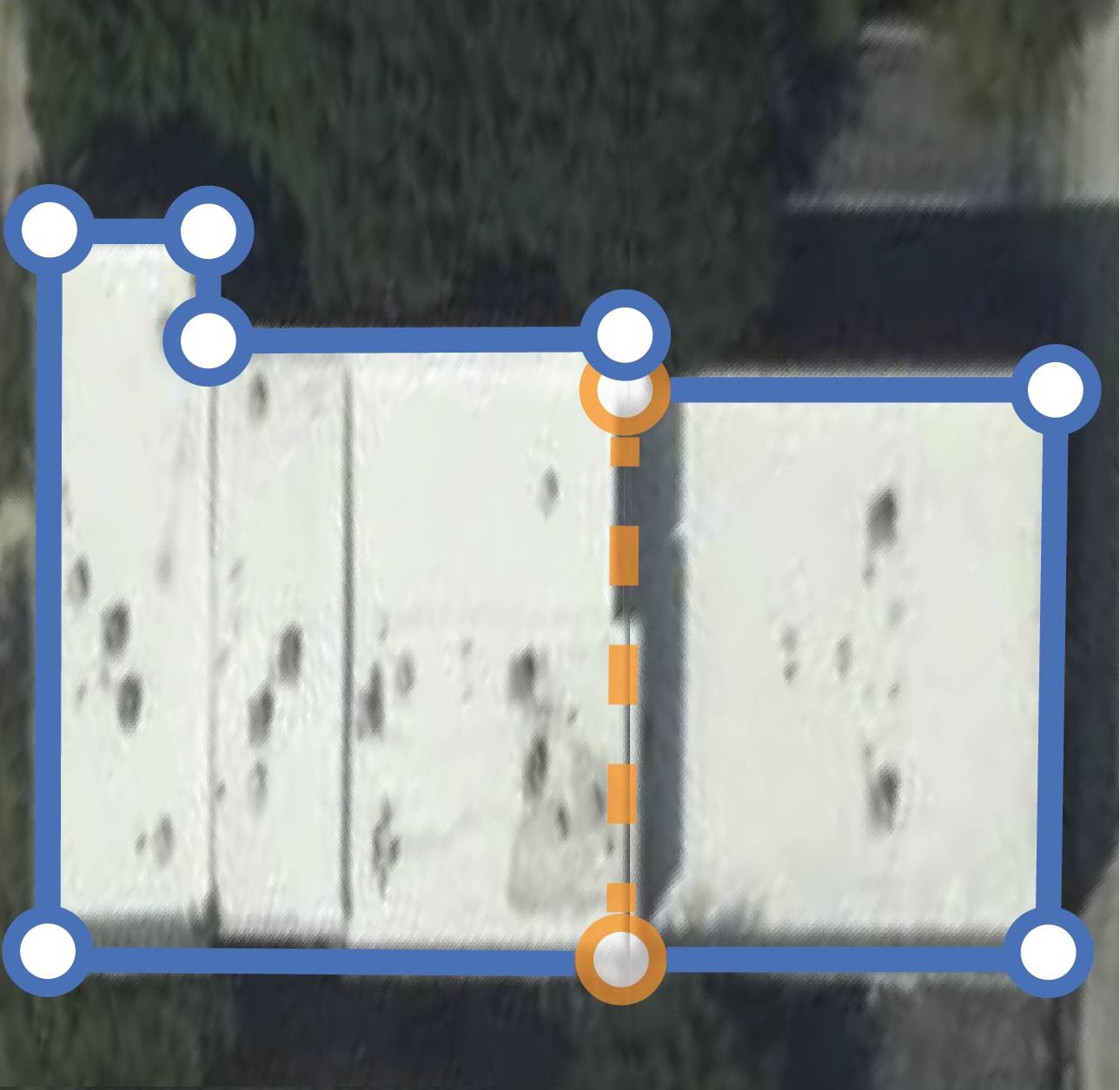}
\end{wrapfigure}
3 out of 10 edges become incorrect.
%(2/2). On the other hand, changes to the junction/edge classifier is small, only (2/9) junctions and (3/10) edges are affected.
%Junction/edge classifiers are more robust to annotation ambiguity,only (2/9) junctions and
%(3/10) edges are affected when orange structure is missing from GT. 
Therefore, we define our region score to be the IoU metric between the union of all the regions against the union of instance segmentations obtained by a pretrained Mask R-CNN~\cite{he2017mask}.

%To solve this issue, we take the union over all local regions, the resulting mask is then used to compute the IoU against the union over all regions instances predicted from a pretrained region Mask R-CNN~\cite{he2017mask}. This IoU score is defined as the region score $C_{\text{region}}$. 

\mysubsubsection{Classification label generation}
For the initial reconstructions and new training samples generated by the exploration, we use the ground-truth to generate classification labels. 
First, we establish corner-to-corner correspondences by greedily matching the closest pair within 7 pixels in distance, where no two corners can match to same GT corner. 
%First, we establish corner-to-corner correspondences between the sample and the ground-truth graph by repeatedly matching the closest pair within 8 pixels while disallowing a match with the same corner twice. 
A junction is labeled as correct if it matches with a GT corner and their incident edge directions are the same with an error tolerance of 10 degrees. An edge is labeled as correct if its end-points match with two corners that are also connected in GT.

\mysubsubsection{Loss function}
The raw output of junction/edge classifiers are pixel-wise scores in the range [-1, 1]. The ground-truth is set to (1, 0, -1) for the pixels over (positive-primitives, background, negative-primitives), where corner and edge primitives are rasterized with 3 pixels of diameters and 2 pixels of thickness, respectively. We enforce pixel-wise Huber loss ($\delta=1.0$) of the output from junction/edge classifiers against the ground-truth. The weight of the loss is set to half for the background pixels.

\begin{figure*}[t]
    \centering
    \includegraphics[width=\linewidth]{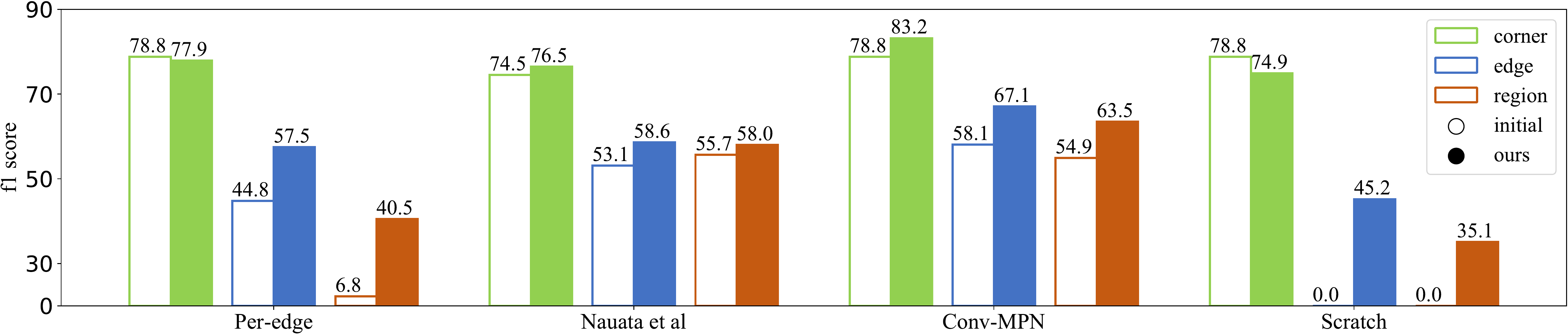}
    %\vspace{-0.4cm}
    \caption{Quantitative evaluations.
    Two baselines (Per-edge and Scratch) and two state-of-the-art algorithms (Nauata \myetal and Conv-MPN) are used to generate initial reconstructions to run our system. 
    %Comparison of our system trained on four different initial graphs against the original initial baselines.
    Corner, edge and region f1 scores are shown in different colors. Empty and solid bars refer to original and our reconstructions. Our system makes consistent improvements in almost all cases.
    %for all cases except for corner metrics in some. %in the edge and region metrics for all the cases.%The corner metrics do not capture the overall reconstruction quality, whereasGreen, blue, red bars represent the c
    }
    %bar is the original initial graph result, solid bar is our system result.} 
    \label{fig:metric}
\end{figure*}

\begin{figure*}
    \centering
    \includegraphics[width=\textwidth]{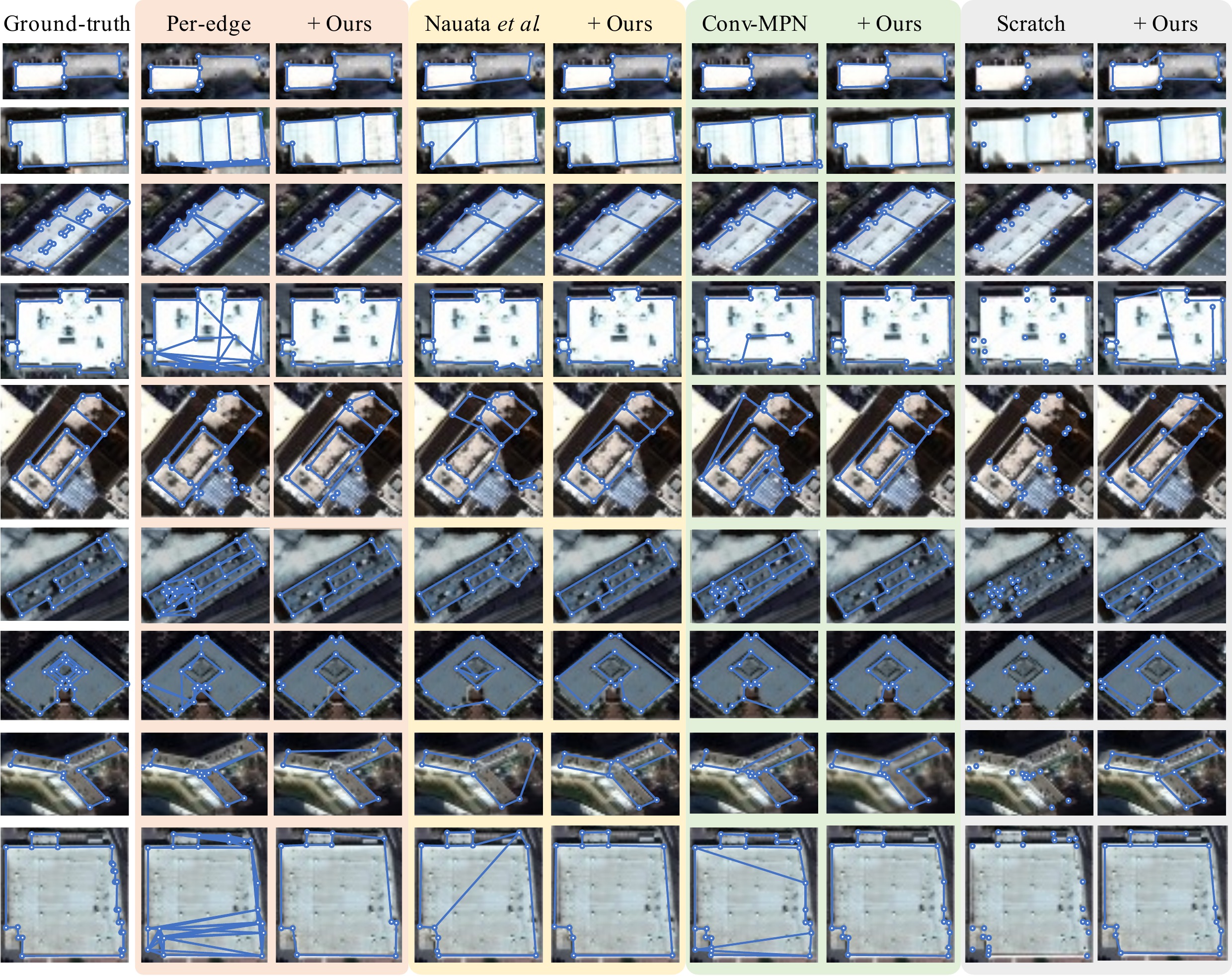}
    %\vspace{-0.4cm}
    \caption{Qualitative evaluations. 
    %Structured graphs generated from our explore-and-classify framework. 
    Improvements are also qualitatively clear in every case.
    %Our system consistently improves the reconstructions in all the cases.
    Refer to the supplementary for more examples.
    %Left most column is the ground-truth structure. Comparison between result from our system and the initial input graph is shown on four different existing algorithms. Our system significantly improves the reconstruction. Additional examples in the supplementary document.
    }
    \label{fig:main_result}
\end{figure*}

\section{Experiments} \label{sec:results}

%\mysubsubsection{Evaluation metrics}
%We follow the evaluation from Nauata \myetal\cite{nauata2020vectorizing}. Corner and edge labelling are described in details in section~\ref{sec:exploration} (classification label generation). We assign detected region to ground-truth region by computing their IoU scores and greedily match the highest pair first. Matched pair with IoU score higher than 0.7 is labelled as correct. We report the final f1 scores based on precision and recall for corner, edge, and region respectively. 

We implement the system in PyTorch~\cite{paszke2019pytorch} and use a workstation with Xeon CPU (20 cores) and dual Titan RTX.
Training is divided into pre-training and fine-tuning phases. In the pre-training phase, initial reconstructions by an existing algorithm are used as training samples and train for 90 epochs. The  fine-tuning phase runs with 2 threads. The first thread conducts exploration while producing more training samples. The second thread fine-tunes the classifiers for 20 epochs.
The networks weights of the classifiers are updated by the second thread after each epoch, which is used by the first thread for the exploration.
%, while updating the network weights of the classifiers after every epoch.
Approximately, 1200 more training samples are generated in each epoch.
%and updates the network weights for one epoch, every after 1200 more training samples are obtained.

%, the training set grows by the exploration process and we launch two threads in parallel, one thread fine-tunes the classifiers for a total of 20 epochs, and another thread explores new graphs. After every fine-tuning epoch, at least 1200 graphs are added to training set and model weights in exploration thread is updated with the latest weights from training thread.
%
%Pre-training is to learn the correctness of initial graph. Training-time exploration launch during fine-tuning stage and train on possible offspring-graphs.

During pre-training, learning rate starts from  $5\times10^{-4}$ and is reduced by half every 30 epochs. During fine-tuning, learning rate starts from $2\times10^{-4}$ and is reduced by half every 5 epochs. In all our experiments, we use the Adam optimizer~\cite{kingma2014adam} with a batch-size of 16. Training takes 8 hours on average.
%an average of 5 hours with an Nvidia RTX 3090. 
During testing, running time depends on the search width and depth.
%we need to evaluate all possible offspring-graphs and the time varies for different search width and depth.
Default setting of width=5, depth=12 takes around 3 to 4 hours for 350 test buildings.

%\yasu{I would say merge 5.1 5.2 and 5.3 into a single subsection, Preliminaries.  Merge 5.4 and 5.5 into Baselines and competing methods. Then to 5.6. 5.7 and 5.8 are good}
%\yasu{We talk about a bit of metrics in the problem review section. We do not need too many lines here. Can you merge 5.1 and 5.2, potentialy 5.3 to put all these details? Feels that 5.6 is way too far to reach while 5.6 is the important piece. reviewers will be dead by the time they read 5.6.}

\subsection{Main results}
\label{sec:main_result}
Our system works with any reconstruction algorithms that produce the initial graphs. We test 2 baselines and 2 state-of-the-art algorithms, where the four experiments are conducted independently without mixing training samples.

\mysubsubsectiona{Per-edge} classifier is the first baseline, which classifies correctness of every edge given building corners from a CNN. See the supplementary document for the details.

\mysubsubsectiona{Nauata \myetal}\cite{nauata2020vectorizing} is the first state-of-the-art algorithm, which relies on integer programming with heuristic objectives. We use the official implementation.

\mysubsubsectiona{Conv-MPN}~\cite{zhang2020conv} is the second state-of-the-art, which use an end-to-end graph neural network to classify the correctness of each edge candidate. We use the official implementation.

\mysubsubsectiona{Scratch} is an extreme case where we only use corners detected by a standard CNN without any edges. A corner detector from the Conv-MPN system is used. 
%Lastly for Scratch, which starts the exploration without any edges,
%When training/evaluating from scratch,
We allow only addition actions in the first 10 steps at training and testing.

%We trained our classifiers to assess the correctness of a given geometry, which can be produced by any existed methods.
%In our experiments, we test four different methods: 1) Conv-MPN\cite{zhang2020conv}, 2) Nauata~\myetal\cite{nauata2020vectorizing}, 3) Per edge classifier and 4) Scratch with detected corners. 

%First two are current state-of-the-art methods in planar graph reconstruction problem, where Conv-MPN uses an end-to-end graph neural network to classify the correctness of each edge jointly and Nauata~\myetal relies on integer programming with heuristic objectives and structural constraints. Per edge classifier is a baseline model that directly performs classification for each edge independently. From scratch is an extreme case that initialized with unconnected corners from Conv-MPN corner detection. 

\vspace{0.1cm}
%\noindent
%For fair comparison, we only train the system with one particular type of initial graph and evaluate on the same type graph (e.g. when evaluating per-edge result, the system is not trained with additional data from other 3 initialization).
We use a simple grid search with the validation set to find the best hyperparameters.
%The hyperparameters based on a simple grid search on the validation set differ slightly.
Scaling weights for the three classification scores (junction, edge, region) in Eq.~\ref{eq:classifier_score}  are
%The scaling weights 
%for the three classifiers ($w_{\hat{j}}, w_{\hat{e}}, w_{\hat{r}}$) in Eq.~\ref{eq:graph_score}
(1, 2, 50) for Per-edge and Conv-MPN, and (1, 1, 50) for Nauata~\myetal and Scratch. The search depth and width are (12, 5) for Nauata~\myetal and Conv-MPN. Per-edge and Scratch requires more exploration and use (20, 5) and (30, 8).
%20, 5) for 
%In our experiments, the optimal junction, edge and region weights are 1,2,50 for both Conv-MPN and per edge, 1,1,50 for both Nauata \myetal and from scratch. The searching depth and width values are 12,5 for both conv-mpn and Nauata \myetal,  20,5 for per edge, and 30,8 for from scratch. 
%fix first 10 steps with addition action set and use full action set after 10 steps. All these hyper-parameters are chosen based on validation set.

%\mysubsubsection{Qualitative result} 
Fig.~\ref{fig:metric} is our quantitative result, comparing corner, edge, and region f1 scores between initial and our reconstructions.
%the final result after passing the initial graph through our system.
The proposed approach consistently improves the edge and the region metrics in every single case.
%Results show that our method consistently improves the reconstruction quality on the edge and region metrics. 
The corner metric degrades in the case of Per-edge and Scratch.
%Corner metric of our system slightly drops for per-edge and scratch. 
However, as shown in Fig.~\ref{fig:main_result}, the corner metric at the lowest level primitive does not capture the overall reconstruction quality.
Original reconstruction by Per-edge and scratch suffer from various artifacts such as dangling edges, overlapping edges, and unlikely structure as a building, all of which are handled well by our system. Even in the extreme case Scratch, our system removes noisy corners and adds meaningful edges to recover most building structure.   
%initial graphs are often broken, e.g. dangling edges, overlap edges, etc, and our system makes significant improvement on those part. This demonstrates that edge and region metrics best reflect our planar graph structure quality. 

While our system does its best in improving original reconstructions, the final quality depends on the initial models. Our system achieves the best performance with Conv-MPN but performs worse with Scratch, which is simply because Scratch requires more actions towards a good reconstruction, making the search space exponentially larger. %This is exacerbated by the fact that we drop the imitation learning and bootstrapping for fair assessment of the core RL modules.

\subsection{Comparison with Reinforcement Learning}\label{rl_setup} 
Reinforcement learning (RL) is an interesting way to tackle structured geometry reconstruction. We compare against two state-of-the-art RL based structured reconstruction systems, Ellis \myetal~\cite{ellis2019write} and Lin \myetal~\cite{lin2020modeling}. 

\mysubsubsection{Algorithm comparison}
Ellis \myetal proposed a neural program synthesis algorithm, one of whose applications is constructive solid geometry reconstruction from occupancy grid. Lin \myetal proposed a system that manipulates a set of primitives to reconstruct a 3d shape. 
%Their implementations are specific to their tasks and cannot be used for fair comparison. 
We adapt the implementation of our system to reproduce their works.
%, while dropping our three specializations.
For both systems, a state is a corner/edge mask image together with the aerial image. An action is represented by the corner/edge mask difference between the current and the next states. The action set is the same as ours. A reward function is based on the entire graph and dependent on the action. We want to assess only the core RL system components and remove the imitation learning in Lin \myetal and the pre-training in Ellis \myetal. The following are more details.

\vspace{0.1cm}
\noindent 
$\bullet$ 
REPL trains a policy network with REINFORCE where reward is 1 for the last state if classification score is within 8.0 from that of the ground-truth, and 0 otherwise.
%it reaches ground-truth and 0 otherwise.
A state value function is also trained based on the reward. 
%They use a $\gamma$ of 1.
%We set a reward of 1 for the last graph if 
We follow their optimization procedure to train the two models with $\gamma$=1 while sampling actions with the policy network.
%Following their work, we use $\gamma$=1, train the two models following based on their optimization procedure, sample actions according to the policy network.
%During training, actions are sampled according to the policy network. 
Sequential Monte Carlo is used at test time.
%finds the best result at test time. 

\noindent
$\bullet$
Lin \myetal originally defines the reward as increase in IoU score between current state and target state. Similarly, we define the reward as an increase in the classification score after each action. Reward discount factor is set to the same 0.9. We train a double DQN~\cite{van2016deep} following their procedure. 

\mysubsubsection{Reduction to RL}
Our system is a special case of temporal difference reinforcement learning (TD-RL) for state value function, whose general formula is 

\begin{eqnarray}
V^{\pi}(s) = \mathbb{E}_{\pi} \left[R(s,a) + \gamma  V^{\pi}(s^\prime)| s\right].
\label{eq:td-v}
\end{eqnarray}
$s$ is the current state and $s^\prime$ is the next state after an action $a$ with reward $R(s,a)$. $\gamma$ and $V$ denote a discount factor and a state value function. $\pi$ is the policy. Our classifiers evaluate goodness of a reconstruction and are equivalent to the value function. Classification labels are based on ground-truth and are equivalent to the reward. Three specializations reduce the TD-RL formulation into our approach: (1) setting $\gamma=0$ to ignore future rewards, (2) making reward solely depends on the state, and (3) Per primitive classifiers design instead of a single function evaluating the goodness of the entire state \cite{ellis2019write, lin2020modeling}. (1) and (2) reduce Eq. \ref{eq:td-v} to $V(s) = R(s)$, where $R(s)$ is the classification labels in our case. (3) breaks reward down to local primitives. We have $V(\tilde{s})=R(\tilde{s})$ for each geometric primitive $\tilde{s}$, which exactly matches our formulation in Sec.~\ref{sec:classifiers}.

\begin{table}[!t]
\caption{Comparison against two state-of-the-art RL-based systems, Ellis \myetal and Lin \myetal based on the corner/edge/region f1 scores. All systems use Conv-MPN reconstructions as the initial models.
%
%We report the f1 scores for corners, edges, and regions. 
\textcolor{orange}{orange} and \textcolor{cyan}{cyan} indicate best and second best scores.}
%\textcolor{magenta}{magenta} colors indicates the top three results of prior free methods.}
\label{rl_baseline}
\centering
\begin{tabular}{lrrr}
\toprule
Algorithm & Corner & Edge & Region\\
\midrule
Conv-MPN~\cite{zhang2020conv} & \textcolor{cyan}{78.8} & \textcolor{cyan}{58.1} & \textcolor{cyan}{54.9}\\
REPL~\cite{ellis2019write} & 72.6 & 43.8 & 15.9 \\
Lin \myetal ~\cite{lin2020modeling} & 74.7 & 51.6 & 35.8\\
+ Ours & \textcolor{orange}{83.2} & \textcolor{orange}{67.1} & \textcolor{orange}{63.5}\\
\bottomrule
\end{tabular}
\end{table}

\begin{figure*}
    \centering
    \includegraphics[width=\textwidth]{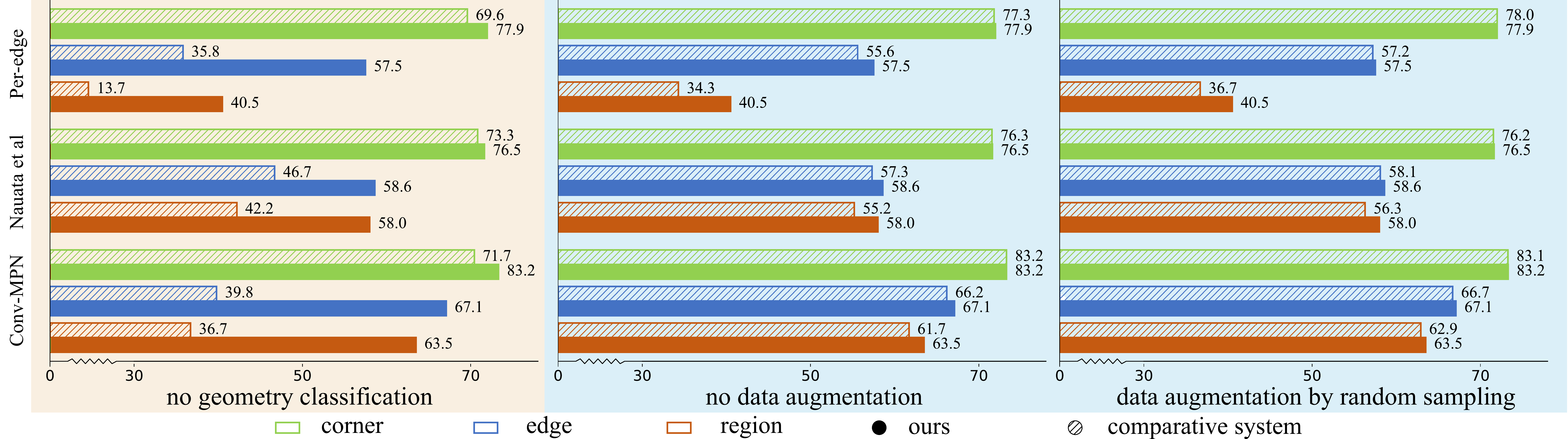}
    %\vspace{-0.4cm}
    \caption{Ablation studies: (Left) The geometry classifier is replaced by a simple heuristic rule based on the corner/edge confidence map (an input to the classifier, see Fig.~\ref{fig:network_structure}). This version does not require any training and merely performs explore-and-classify by heuristic rules. (Right) Data augmentation by the exploration module is replaced by no augmentation or random sampling. Each study reports numbers  with three different initial reconstruction algorithms.
    }
    \label{fig:ablation}
\end{figure*}

%\subsection{RL comparison}
%In Sec.~\ref{section:rl}, we discuss the relationship of our model to reinforcement learning and discuss adaption of the two existing systems, REPL~\cite{ellis2019write} and Lin \myetal ~\cite{lin2020modeling}, to the planar graph reconstruction task.
%We trained the two RL models on Conv-MPN initial graph and results are shown in 

\mysubsubsection{Empirical comparisons}
Table~\ref{rl_baseline} is the quantitative comparison of two RL based methods with ours, where Conv-MPN is the initial reconstructions for all the three systems.
It is a surprise that Ellis \myetal and Lin \myetal rather make the results worse from the original Conv-MPN reconstructions. This shows the limitation of using RL on the structured data, which fails to learn patterns in architectural geometry and rather become a noisy guidance damaging the reconstruction.

%
%In fact, both REPL and Lin \myetal are worse than original Conv-MPN baseline, with REPL trained with sparse reward having the lowest f1 scores. This proves that a single global value function trained with discounted future rewards can not effectively solve our reconstruction task. 

\begin{figure}[!t]
    \centering
    \includegraphics[width=\linewidth]{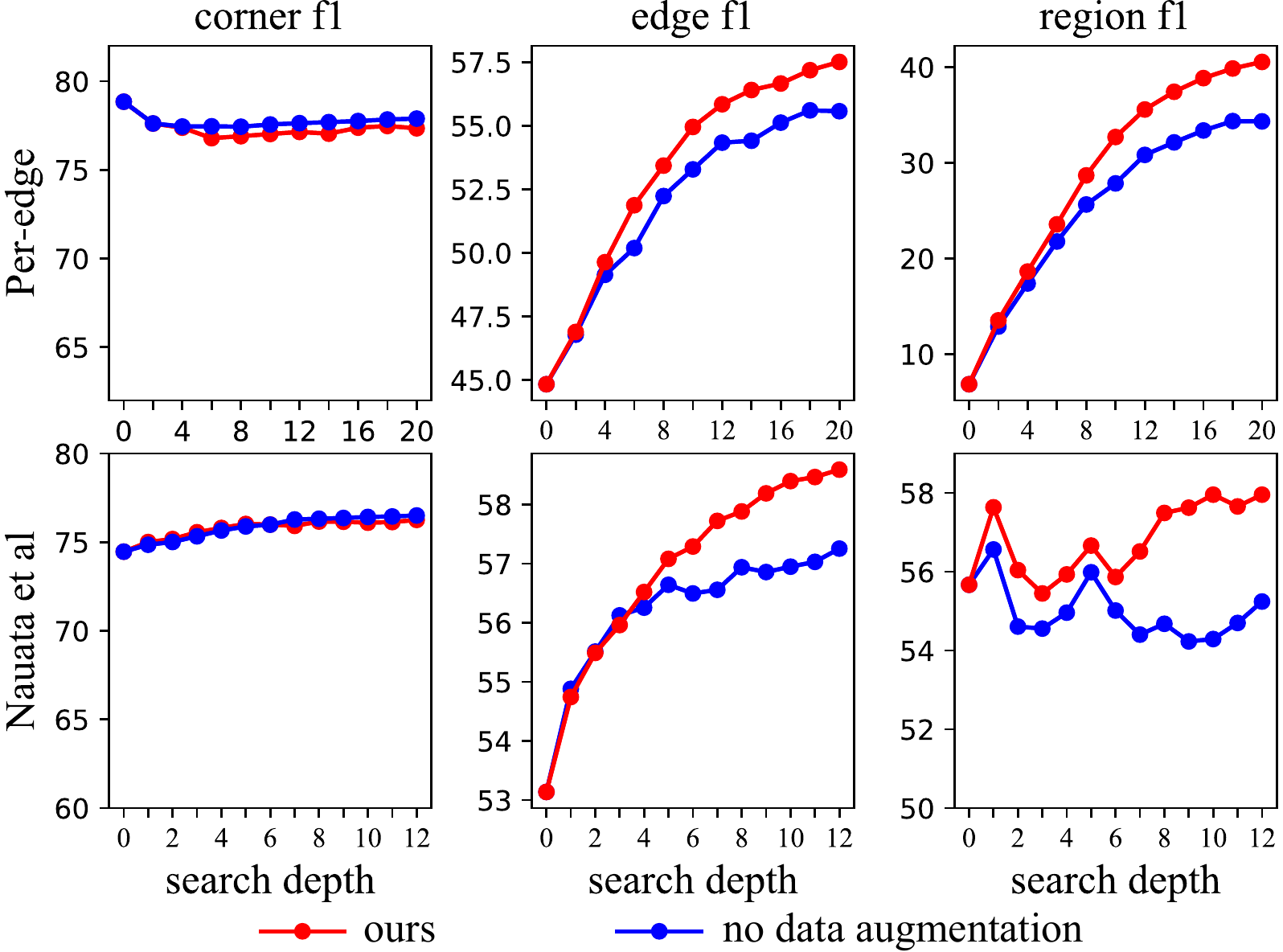}
    %\vspace{-0.5cm}
    \caption{Ablation study on the effectiveness of exploration module. We show the performance gap between ours and \textit{no data augmentation}. The top (resp. bottom) row shows the corner/edge/region f1 scores when using Per-edge (resp. Nauata \myetal) as the initial reconstructions.
    %
    %Performance-search depth curve for comparison between ours and ours without exploration on Per-edge and Nauata \etal initial graphs. Performance gap for edge and region f1 scores increase with larger search depth.
    The performance gap grows as search depth increases, because data augmentation becomes more effective as reconstruction samples differ more from the originals.
    }
    \label{fig:curve}
\end{figure}

\subsection{Ablation studies}
\label{sec:ablation}
We verify the contributions of each individual components in our system together with the assessment on the generalization capability.

\mysubsubsection{Exploration module}
%Effect of geometry exploration}
Our exploration module collects training data dynamically by utilizing the classification module. We experiment two ablation approaches for the training data collection. The first one denoted as \textit{no data augmentation}, which does not explore and simply use the initial reconstructions as the only training data.
%We compare against two methods that do not use training-time exploration. The first method simply doesn't explore data, i.e. only trained on initial graph, and is denoted as ``w/o exp''. 
The second one denoted as \textit{data augmentation by random sampling}, which uses random sampling instead of our classifiers to perform exploration and select training samples.
%strategy during training, i.e. augment dataset by randomly choosing the reconstruction candidates without relying on the classification score. 
Fig.~\ref{fig:ablation} shows that both two ablation approaches are inferior to our exploration module with an exception of one entry in the corner metric with Per-edge initial reconstruction, which is not a good reflection of overall reconstruction quality as discussed in Sec.~\ref{sec:main_result}. For both \textit{no data augmentation} and \textit{data augmentation by random sampling}, the region metric gap with ours is larger when initial reconstructions are of low quality (i.e. Per-edge).

Fig.~\ref{fig:curve} further shows that the performance gap between \textit{no data augmentation} and our exploration module grows as the search depth increases. This agrees with our intuition that the data augmentation is more effective when building models differ more from the initial reconstructions.

\mysubsubsection{Classification module}
%Effect of geometry classification}
To verify the effectiveness of the junction/edge classifiers,
%cation (junction/edge),
we replace them with a heuristic method based on the corner/edge confidence images utilized inside the classifiers (See Fig.~\ref{fig:network_structure}).
Concretely, we simply use the corner/edge confidence images for pooling, instead of the pixel-wise classification scores. The rest of the pipeline stays the same.
%
%generated by a pre-trained CNN
%we replace our trained classifiers with the corner/edge confidence maps and use this to compute the classification score. All other settings such as beam search are the same.
Fig.~\ref{fig:ablation} shows that the lack of classification module significantly degrades the performance, in fact, made the results even worse than the initial reconstructions. %This agrees with our observation in Table~\ref{rl_baseline}, where poor classifiers rather become a noisy guidance damaging the reconstruction.

%We denote this comparative method as \textbf{w/o classifier}. Table \ref{table_ablation} shows that our system using the trained classifiers is clearly better than ``w/o classifier'' on all three types of initial graphs. In particular, the performance for 

%"w/o classifiers" is worse than most original results, especially for edge and region metrics. This indicates that our geometry classifiers can successfully inference the correct geometric topology.

\mysubsubsection{Search strategy}
%Fig.~\ref{fig:depth_curve} shows our system improve reconstructions over multiple iterations of search. Different initial graphs has different optimal setting. Generally, a deeper search scheme produces better reconstruction results until system reaches the performance limit, for example, \fuyang{describe one curve of the plot}. On the other hand, a deep search means it costs more time to search the best result. The deepest test in our experiment is from scratch experiment: width=8, depth=30. It takes two days to finish 350 testing data.
%Beam search is not the only reasonable search algorithm. 
Sequential Monte Carlo~\cite{doucet2001smc} (SMC) is another popular search method. 
We replace Beam search with SMC at test time, sampling 5 graphs based on the classification scores at each iteration and repeat for 12 times (our default search depth).
%from all the off-springs
%at inference time instead of the Beam search. SMC is a set of methods used for sampling a sequence of latent variables conditioned on observed variables. In our problem, we conduct SMC in the following way: sample 5 graphs from all the off-springs weighted by classification score and repeat 12 times.
%
Table~\ref{table_search} compares SMC against Beam search while using Conv-MPN as the initial reconstructions. Our system achieves clear improvements over the original reconstructions in both cases.
%compares SMC with beam search usuing Conv-MPN result as initial graph. We see that beam search achieved better corner and edge performance. Region score is slightly lower than SMC but still much better than original Conv-MPN result. This proves that our system is effective with multiple search strategies.

%\begin{table}[H]
%\caption{Ablation study for different search algorithms}
%\label{table_matrix}
%\centering
%\begin{tabular}{l|rrr}
%\toprule
%\diagbox{Train}{Test} & Conv-MPN & Nauata \etal & Per-edge\\
%\hline
%Conv-MPN & 83.2/67.1/63.5 & 76.6/58.5/57.7 & 81.1/60.3/35.8\\
%Nauata \etal & 83.1/67.0/63.0 & 76.5/58.6/58.0 & 80.5/60.1/36.5 \\
%Per-edge & 82.7/65.4/52.4 & 75.0/56.0/51.0 & 77.9/57.5/40.5 \\
%\hline
%\end{tabular}
%\end{table}

%\begin{figure}[!t]
%    \centering
%    \includegraphics[width=\linewidth]{figures/matrix.png}
    %\vspace{-0.4cm}
%    \caption{Result for training and testing on different reconstruction graphs. [A] is graphs from Conv-MPN, [B] is from Nauata \myetal, [C] is from Per-edge classifier. Left is corner f1 score, and right is edge f1 score.} 
%    \label{fig:matrix}
%\end{figure}

\mysubsubsection{Generalization capability}
We evaluate the generalization capability of our approach across different initial reconstructions based on the corner/edge f1 scores in Table.~\ref{fig:matrix}.
For example, the value of \textcolor{orange}{76.6} in the second column indicates a score when Conv-MPN~\cite{zhang2020conv} is used at training and Nauata \myetal~\cite{nauata2020vectorizing} is used at testing.
Our system is able to achieve high performance regardless of the training/testing data combination.
%we train our system with Conv-MPN reconstructions, and test on Per-edge. 
%
%We verify the generalization ability of our classifiers by training and testing on different type of graphs (e.g. train on Conv-MPN, test on per-edge). Fig.~\ref{fig:matrix} shows the result on three types of reconstructions. We report the corner and edge scores since we only train classifiers on those two primitives. 
%
One interesting discovery is the right most column where Per-edge reconstructions are used at test-time. It is a surprise that using Conv-MPN or Nauata \myetal at training is better. 
%Usually, it is better to make the training and testing configurations the same.
Our hypothesis is that
%The classifiers learn to evaluate local geometric primitives. 
reconstructions by Conv-MPN or Nauata \myetal contain both high quality local geometries (close to GT) as well as poor local geometries (i.e., reconstruction failures), enabling the classifiers to learn from a diverse set of training samples. On the other hand, Per-edge reconstructions are mostly poor as indicated by the original f1 scores, making it difficult for the classifiers to learn diverse scenarios.

%when testing on Per-edge models, 
%Conv-MPN at training is the best  regardless of test-time initial reconstructions. 
%out to be the best regardless of the type of test-time initial reconstructions.

%Results show that classifiers trained on Conv-MPN and Nauata \myetal have very good generalization ability. Performance is no different from the best diagonal values when tested on Nauata \myetal and Conv-MPN. Per-edge scores are even better than the diagonal values, which is expected since per-edge contains many noisy information (i.e. self-intersection). Classifiers trained on it will have worse performance. Still, model trained on per-edge has significantly better scores than the original when tested on the other two types. This verifies the generalization ability of our classifiers on different types of reconstructions and their ability to classify correctness of local primitives even when trained with noisy data. 

\begin{table}[tb]
\caption{Our system works well with different search algorithms.}
\label{table_search}
\centering
\begin{tabular}{l|rrr}
\toprule
Algorithm & Corner & Edge & Region\\
\midrule
Beam search & \textbf{83.2} & \textbf{67.1} & 63.5\\
SMC~\cite{doucet2001smc} & 81.2 & 65.6 & \textbf{64.0}\\
\bottomrule
\end{tabular}
\end{table}

\begin{table}[tb]
\caption{Assessing generalization capability across different initial reconstructions. 
%Result for training and testing on different reconstruction graphs. 
Columns and rows indicate the initial reconstructions at testing and training time, respectively. (corner/edge) f1 scores are reported in the table. The third row (labeled Original) shows the f1 scores of the original reconstructions by the three systems as a reference.}
\label{fig:matrix}
\centering
\begin{tabular}{l|ccc}
\toprule
& \multicolumn{3}{c}{Test-time initialization}\\
 \cmidrule{2-4} 
 & Conv-MPN & Nauata \myetal & Per-edge\\
\midrule
Original & 78.8 / 58.1 &74.5 / 53.1 &  78.8 / 44.8\\
\midrule
Conv-MPN & \textcolor{orange}{83.2} / \textcolor{orange}{67.1}  &\textcolor{orange}{76.6} / \textcolor{cyan}{58.5}  &\textcolor{orange}{81.1} / \textcolor{orange}{60.3} \\
Nauata \myetal &\textcolor{cyan}{83.1} / \textcolor{cyan}{67.0}  &\textcolor{cyan}{76.5} / \textcolor{orange}{58.6}  &\textcolor{cyan}{80.5} / \textcolor{cyan}{60.1} \\
Per-edge &82.7 / 65.4  &75.0 / 56.0  &77.9 / 57.5 \\
\bottomrule
\end{tabular}
\end{table}

%\renewcommand{\arraystretch}{1.1}
%\begin{table*}[!t]
%\caption{Evaluation: Different initial graph. \fuyang{explain why scratch is 0 for edge/region. Replaced by Fig \ref{fig:metric}}}
%\label{table_initial}
%\centering
%\begin{tabular}{lcccccccc}
%\toprule
%& \multicolumn{2}{c}{Per-edge} & \multicolumn{2}{c}{IP} & \multicolumn{2}{c}{Conv-MPN} & \multicolumn{2}{c}{From-scratch}\\
%\cmidrule(lr){2-3}\cmidrule(lr){4-5}\cmidrule(lr){6-7}\cmidrule(lr){8-9}
%& Original & Ours & Original &  Ours & Original &  Ours  & Original & Ours\\
%\midrule
%Corner & 78.8 &  77.9 &  74.5 &  76.5 &  78.8 &  83.2 &  78.8 &74.9 \\
%Edge   & 44.8 &  57.5 &  53.1 &  58.6 &  58.1 &  67.1 &  0 &45.2  \\
%Region & 6.8  &  40.5 &  55.7 &  58.0 &  54.9 &  63.5 &  0 &35.1  \\
%\bottomrule
%\end{tabular}
%\end{table*}

%\begin{table}[!t]
%\caption{Evaluation: Different initial graph.}
%\label{table_initial_short}
%\centering
%\begin{tabular}{lcccc}
%\toprule
%& \multicolumn{2}{c}{Conv-MPN} & \multicolumn{2}{c}{IP} \\
%\cmidrule(lr){2-3}\cmidrule(lr){4-5}
%& Original & Ours & Original &  Ours\\
%\midrule
%Corner & 78.8 & 83.2 & 74.9 & 77.7  \\
%Edge   & 58.1 & 67.1 & 53.4 & 59.2  \\
%Region & 54.5 & 63.3 & 57.3 & 58.2  \\
%& \multicolumn{2}{c}{Per-Edge} & \multicolumn{2}{c}{Scratch}\\
%\cmidrule(lr){2-3}\cmidrule(lr){4-5}
%& Original & Ours & Original &  Ours\\
%\midrule
%Corner & 79.0 & 77.9 & 78.8 & 75.6\\
%Edge   & 45.0 & 55.3 & 0 & 46.1 \\
%Region & 7.2  & 32.5 & 0 & 36.3 \\
%\bottomrule
%\end{tabular}
%\end{table}

\begin{figure}[t]
    \centering
    \includegraphics[width=\linewidth]{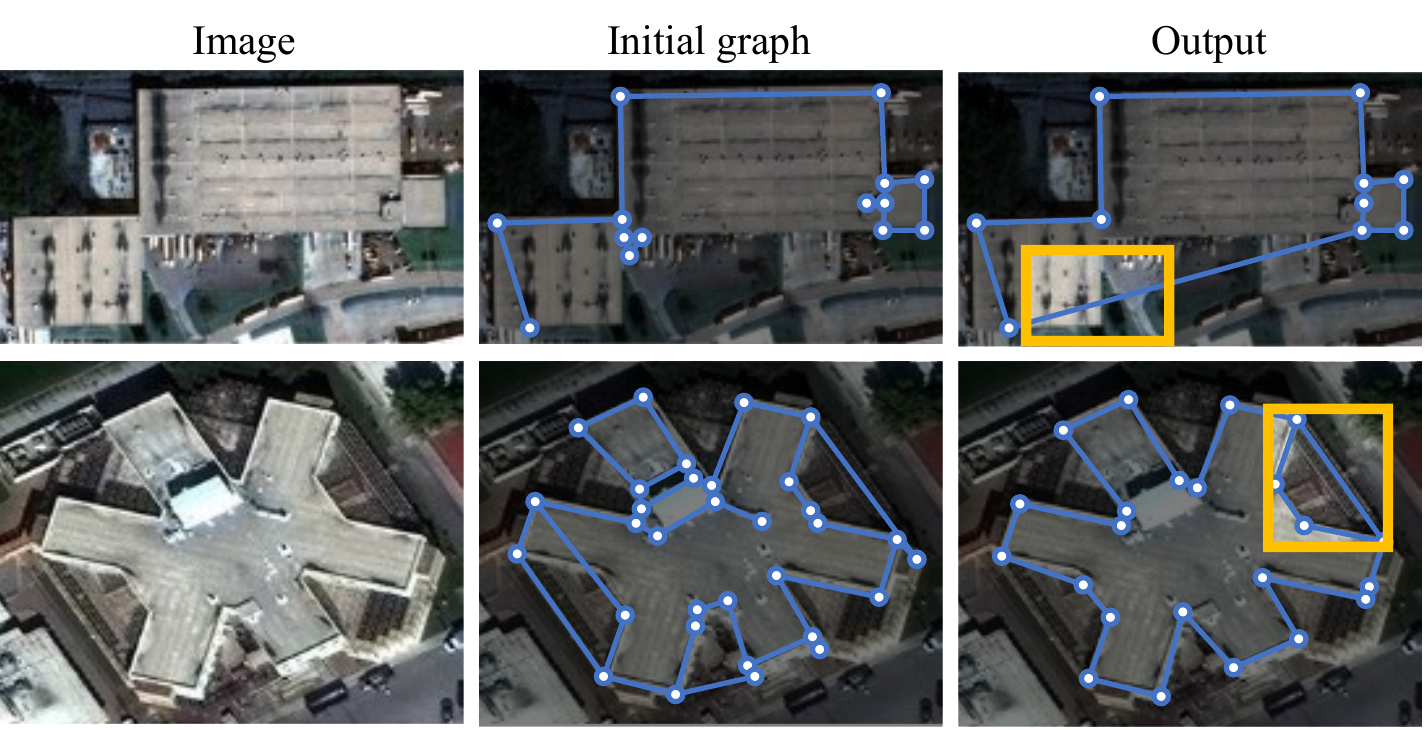}
    %\vspace{-0.4cm}
    \caption{Failure modes. The top row shows an example suffering from missing corners. The bottom row shows a complex building with weak image signal. 
    %Yellow region outlines the incorrect classification made by our system. } 
    }
    \label{fig:failure}
\end{figure}

\subsection{Limitations}
While the proposed system achieves the best performance over the competing state-of-the-art with a clear margin, there are a few limitations. First, test-time inference is slow, which is a general limitation for any search based methods. Next, The system also has a few major failure modes (See Fig.~\ref{fig:failure}). The first mode comes from missing corners. Corner addition is a challenging action with many degrees of freedom. Our system fails to recover from extreme corner detection failures, e.g. a region misses multiple corners. The second mode comes from the classification mistake when the image signal is weak. Our CNN-based classifiers are inferior to the human vision system in analyzing the consistency of structured geometry and an image.

%occurs when the image sinnal is weak.

%Our system is slower than one-shot regression methods (e.g. Conv-MPN) at test time due to the use of beam search. 
%Two failure cases are also illustrated in Fig.~\ref{fig:failure}. Our method can not recover missing corners not included in the adding action set, and it fails on hard data, such as when image signal is weak.
\section{Conclusion}
This paper presents a novel explore-and-classify framework for structured outdoor architecture reconstruction, seeking to improve the quality of imperfect building reconstructions. Our system learns to classify the correctness of primitives while exploring the space of reconstructions via heuristic actions. Qualitative and quantitative evaluations %verify the effectiveness of our idea and
demonstrates significant improvements over all state-of-the-art methods. 
%Our system also addresses missing corners in some special cases, but still can't solve all of them, which we left it for the future work.
%The current popular approaches or is to structured reconstruction are to infer the structure in ``one-shot'' regression which is too difficult to encode the general geometric shape priors. On the contrary, our system encodes such priors by learning to verify the graph correctness. 
While we verify the effectiveness of our idea on one particular reconstruction task, the computational framework is general. We believe that the ideas are potentially applicable to many other structured reconstruction tasks and beyond.
%s a potential to improve the performance of state-of-the-art algorithms on broader 
%open a new line for structured geometry reconstruction.
%Future work includes solving the missing corner issue and designing a local region classifier robust to annotation ambiguity. %We will share our code and data to promote further research.

\mysubsubsection{Acknowledgement}
This research is partially supported by NSERC Discovery Grants with Accelerator Supplements and DND/NSERC Discovery Grant Supplement.

\clearpage

{\small
\bibliographystyle{ieee_fullname}
\bibliography{egbib}
}

\end{document}